\documentclass[runningheads]{llncs}
\usepackage{graphicx}
\usepackage{booktabs}
\usepackage{amsmath}
\usepackage{amssymb}
\usepackage{multirow}
\usepackage{xcolor}
\usepackage{algorithm}
\usepackage{algpseudocode}
\usepackage{enumitem}
\usepackage{orcidlink}  

\begin{document}

\title{Training-Free Logical and Structural Anomaly Detection via Calibrated Fusion}
\titlerunning{Calibrated Fusion of Frozen Cues for Anomaly Detection}

\author{Changyi Li\inst{1}\thanks{Corresponding author.}\orcidlink{0009-0004-9034-9122} \and
Miao Yu\inst{2}\orcidlink{0000-0002-9560-1073} \and
Kai Dong\inst{1}\orcidlink{0000-0002-5873-8054} \and
Yu Xiao\inst{1}\orcidlink{0000-0002-4517-3779}}
\authorrunning{C. Li et al.}
\institute{Aalto University, 02150 Espoo, Finland\\
\email{\{changyi.li, kai.dong, yu.xiao\}@aalto.fi}
\and
Harbin Engineering University, 150001 Harbin, China\\
\email{miao.yu@hrbeu.edu.cn}}

\maketitle

\begin{abstract}
Industrial anomaly detection must handle two distinct defect families: structural anomalies, which manifest as local texture corruptions, and logical anomalies, which violate global rules on object count, composition, or arrangement. Existing detectors typically favor one family at the expense of the other. In particular, training-free methods effectively exploit frozen representations but lack an explicit notion of object count, while methods that reason about counts usually rely on category-specific component modeling. We show that counting ability can be introduced into training-free anomaly detection without additional training or part-level supervision. Our key idea is a normal-set calibration that aligns heterogeneous anomaly cues using statistics from normal images, enabling their direct fusion within a unified training-free framework. Built upon this calibration, our detector combines complementary frozen cues to address both logical and structural anomalies. On MVTec-LOCO, our method achieves image-level AUROCs of $89.0$ and $95.9$ on logical and structural anomalies, respectively, yielding a $92.5$ average—the best among training-free detectors in our comparison. It remains competitive with methods requiring network training or part annotations, while its structural variant matches PatchCore on MVTec-AD ($99.1$ image-AUROC), suggesting that the proposed calibration generalizes beyond logical anomaly detection.

\end{abstract}

\begin{figure}[t]
\centering
\includegraphics[width=0.85\textwidth]{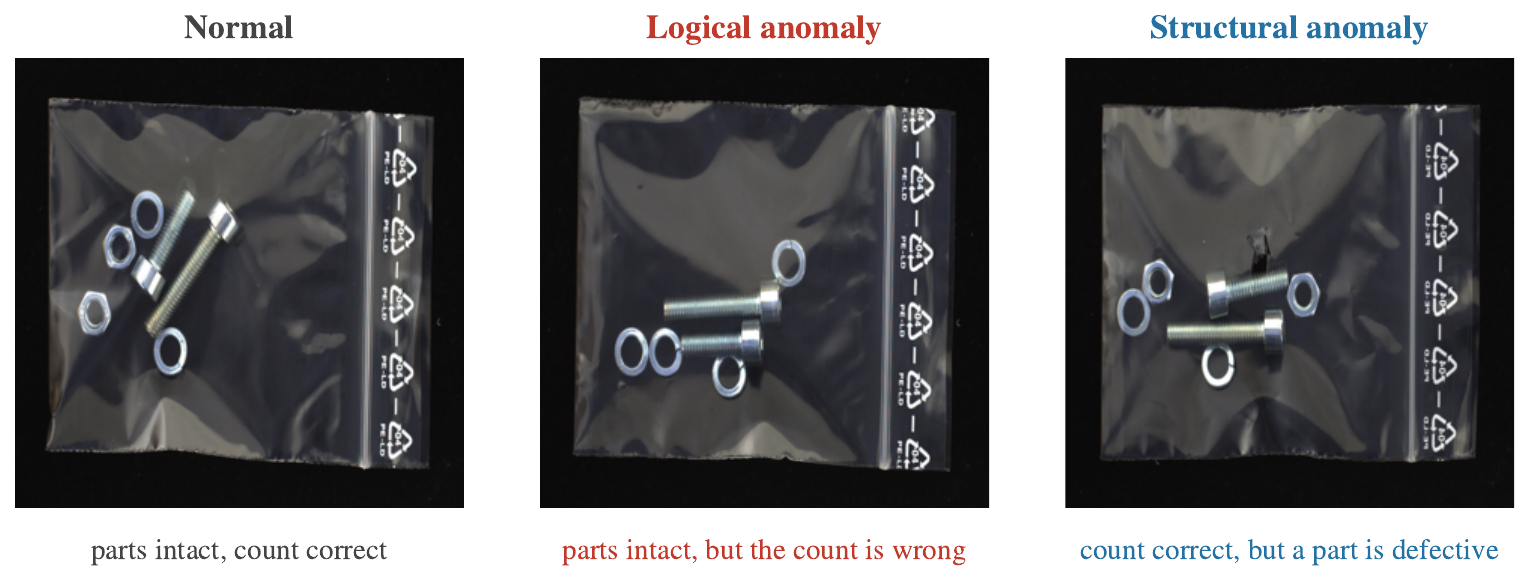}
\caption{
Logical and structural anomalies on the same product (left: normal reference). A logical anomaly (middle) preserves local appearance but violates global rules on count or composition, whereas a structural anomaly (right) preserves global composition but disrupts local appearance.
Handling both requires complementary global and local reasoning, yet existing detectors often excel at one at the expense of the other.
}
\label{fig:teaser}
\end{figure}

\section{Introduction}

Industrial visual inspection aims to detect anomalies using only normal images for training. The MVTec-LOCO benchmark~\cite{bergmann2022loco} highlighted an important distinction between two defect families. Structural anomalies are local appearance corruptions such as scratches, contamination, or deformation, whereas logical anomalies preserve local appearance but violate global rules, for example, through an incorrect number of parts, missing or swapped components, or invalid arrangements. A screw bag with too few nuts contains only normal-looking nuts, so no local patch is anomalous, yet the image is defective (Fig.~\ref{fig:teaser}). Detecting both families within a single model is challenging because they require opposite inductive biases: structural anomalies rely on local evidence, whereas logical anomalies require reasoning over global count and composition.

Existing approaches expose a tension between counting ability and ease of deployment. Training-free methods reuse frozen representations and avoid category-specific optimization. Representative examples such as SINBAD~\cite{cohen2023sinbad} perform well on logical anomalies by aggregating element sets into permutation-invariant descriptors, but they lack an explicit notion of object count and therefore struggle when anomalies arise solely from changes in the number of otherwise normal parts. In contrast, methods such as PSAD~\cite{kim2024psad} and CSAD~\cite{hsieh2024csad} explicitly count components using category-specific part segmenters, achieving strong logical performance at the cost of additional training and annotation. This trade-off undermines the training-free property that makes industrial inspection systems attractive in practice.

This motivates a simple question: can a training-free detector acquire explicit counting ability without per-category training? Our key idea is a normal-set calibration that places heterogeneous anomaly cues extracted from frozen models onto a common scale, enabling them to be aggregated within a unified training-free framework. Building on this calibration, we combine complementary appearance and counting cues to address both structural and logical anomalies, while a subsequent $p$-norm soft-OR preserves strong evidence from individual branches rather than averaging it away.

A diagnostic oracle analysis suggests that signal availability is not the bottleneck: across MVTec-LOCO categories, at least one branch correctly identifies nearly every anomaly. The primary challenge therefore lies in effective aggregation, and the calibrated fusion substantially narrows the gap between naïve combinations and the oracle upper bound. On MVTec-LOCO, our method achieves the best average image-level AUROC among training-free detectors ($92.5$) and the strongest structural performance in our comparison, while remaining competitive with methods that require network training or part-level annotations.

Our contributions are threefold. We introduce a unified training-free framework that addresses both logical and structural anomalies without gradient-based optimization, large language models, or category-specific models. The framework relies on a normal-set calibration and a $p$-norm soft-OR to render heterogeneous frozen cues directly fusible without suppressing single-branch evidence, and further incorporates open-vocabulary counting to provide competitive logical reasoning together with interpretable evidence attribution.

\section{Related Work}

\noindent \textit{Structural anomaly detection.} Most industrial anomaly detectors target structural defects through local appearance modeling, including memory-based, one-class, and reconstruction-based approaches~\cite{roth2022patchcore,batzner2024efficientad,yang2023slsg,li2025s2tkd}. While highly effective on texture defects, their locality makes them ineffective for logical anomalies, where every local region may appear normal despite a globally incorrect composition.

\noindent \textit{Logical anomaly detection.} Existing logical anomaly detectors broadly follow two directions. Global approaches such as GCAD~\cite{bergmann2022loco},  SINBAD~\cite{cohen2023sinbad}, PUAD~\cite{sugawara2024puad}, and UNISLAD~\cite{li2026unislad} capture composition through holistic representations, but lack an explicit notion of object count and therefore remain weak when anomalies arise solely from changes in the number of otherwise normal parts. Counting-based approaches such as ComAD~\cite{liu2023comad}, CSAD~\cite{hsieh2024csad}, and PSAD~\cite{kim2024psad} explicitly reason about object counts through component modeling, achieving strong logical performance at the cost of category-specific clustering, segmentation training, or part annotations. Existing logical detectors therefore expose a clear trade-off: they are either training-free without explicit counting, or counting-aware at the expense of per-category modeling.

\noindent \textit{Foundation models for anomaly detection.} Foundation models provide transferable primitives for anomaly detection. The work most closely related to ours, SAM-LAD~\cite{peng2025samlad}, detects logical anomalies by establishing correspondences between segmented objects in query and reference images. In contrast, we use a frozen promptable segmenter as an open-vocabulary counting primitive, allowing counting signals to be incorporated without object matching or additional training.

\noindent \textit{Fusion of anomaly cues.} Combining multiple anomaly cues is common in industrial anomaly detection. Existing methods typically aggregate normalized scores through fixed-weight sums or averages~\cite{jeong2023winclip}, implicitly assuming that heterogeneous cues are already comparable. In practice, frozen-model branches often produce scores with substantially different scales and distributions, making naïve fusion ineffective. Our normal-set calibration instead rescales each branch using statistics computed solely from normal images, enabling heterogeneous cues to be fused directly without learned fusion weights.

\noindent Taken together, existing methods either remain training-free but lack explicit counting or obtain counting through category-specific component models. Our method bridges this gap by introducing a training-free counting view from a frozen promptable model and unifying heterogeneous anomaly cues through normal-set calibration, enabling a single detector to address both logical and structural anomalies.

\section{Method}
\label{sec:method}

\subsection{Overview and notation}

Given normal training images from a single category, our goal is to assign an anomaly score to a test image $x$. Following Fig.~\ref{fig:arch} from left to right, the detector extracts complementary anomaly cues spanning appearance and counting using five frozen branches, $\mathcal{B}=\{\textsc{deep},\textsc{pix},\textsc{pc},\textsc{pc-cnn},\textsc{sam}\}$
Four branches capture appearance-based evidence through set features and patch memories, while the fifth branch, \textsc{sam}, provides an open-vocabulary counting cue for logical anomalies. Each branch outputs a raw anomaly score $s_b$.

Although these branches provide complementary signals, their raw scores follow different distributions and cannot be fused reliably. Our key idea is a normal-set calibration that rescales each raw score $s_b$ into a calibrated score $\hat{s}_b$ (Sec.~\ref{sec:calib}), placing heterogeneous cues onto a common scale using only normal images. The calibrated scores are then aggregated through a $p$-norm soft-OR to produce the final anomaly score $S$ (Sec.~\ref{sec:fusion}), preserving strong evidence from individual branches rather than averaging it away.

Beyond anomaly scoring, the same calibrated scores further provide interpretable evidence attribution (Sec.~\ref{sec:explain}), identifying whether a detection is primarily supported by logical or structural cues and, when applicable, reporting the corresponding count rationale. All components remain frozen and only store or summarize normal images from the target category. 

\begin{figure}[t]
\centering
\includegraphics[width=\textwidth]{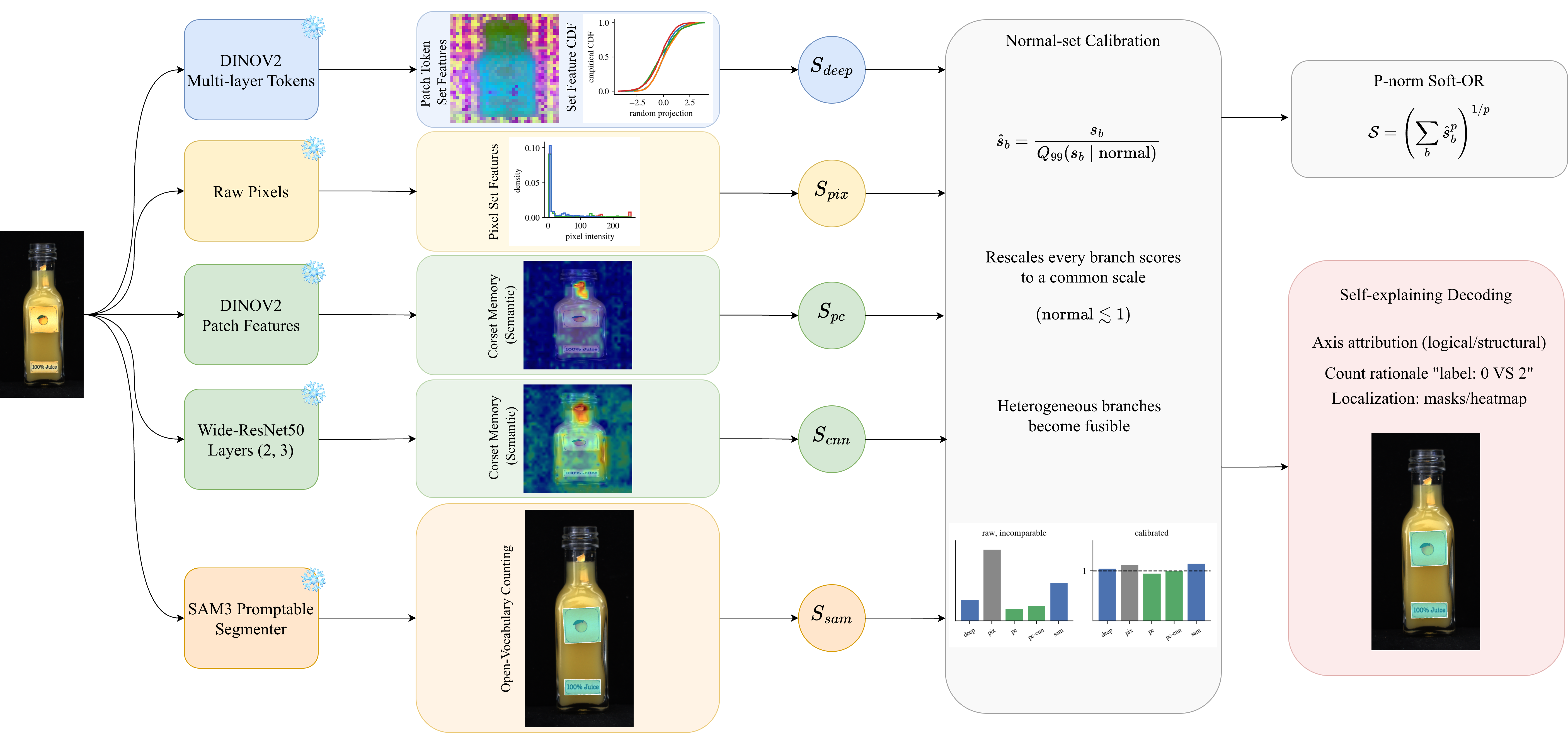}
\caption{
Overview of the proposed calibrated fusion framework. Frozen branches extract complementary anomaly cues spanning appearance, texture, and open-vocabulary counting. Our key idea, normal-set calibration, rescales these heterogeneous scores onto a common scale using only normal-image statistics, enabling their fusion through a $p$-norm soft-OR into the final anomaly score $S$. The calibrated scores also yield interpretable evidence attribution for logical and structural decisions.
}
\label{fig:arch}
\end{figure}

\subsection{Appearance representations}

Appearance evidence in our framework serves two complementary roles: capturing global composition and detecting local structural defects.

\noindent \textit{Set features.} Following SINBAD~\cite{cohen2023sinbad}, the first appearance representation describes an image by the distribution of its elements rather than by their spatial layout, making it sensitive to composition while remaining invariant to exact position. We represent an image $x$ as a set of elements
$E(x)=\{e_i\}_{i=1}^{N}$, where $e_i$ is either a frozen DINOv2~\cite{oquab2024dinov2} token extracted from multiple depths for the semantic branch \textsc{deep}, or a raw RGB patch for the low-level branch \textsc{pix}. Following SINBAD, this distribution is summarized through a permutation-invariant descriptor. Specifically, given random projection directions $\{u_k\}_{k=1}^{K}$ and quantile levels $\{\theta_{k,l}\}$, we compute
\begin{equation}
\phi_{k,l}(x)=\frac{1}{N}\sum_{i=1}^{N}\mathbf{1}\!\left[\,u_k^{\top}e_i\le\theta_{k,l}\,\right],
\end{equation}
which stack into a descriptor $\phi(x)\in\mathbb{R}^{KL}$
The resulting descriptors are whitened and scored by nearest-neighbour comparison against normal descriptors, yielding branch scores $s_{\textsc{deep}}$ and $s_{\textsc{pix}}$. Because $\phi$ encodes \emph{how often}, rather than \emph{where}, patterns occur, it captures global composition while remaining insensitive to exact spatial arrangement.

\noindent \textit{Dual patch memory.} Global composition alone is insufficient for structural anomalies, which often manifest as subtle local defects. We therefore maintain two complementary PatchCore~\cite{roth2022patchcore} memories. Each memory stores a coreset $M$ of normal patch features and scores an image by the average distance of its most anomalous patches to the memory:
\begin{equation}
s_{\textsc{pc}}(x)=\frac{1}{|\mathcal{T}|}\sum_{r\in\mathcal{T}}\;\min_{m\in M}\lVert f_r-m\rVert_2,
\end{equation}
where $\{f_r\}$ are the patch features of $x$, and $\mathcal{T}$ contains the top $1\%$ of patches ranked by nearest-memory distance. The semantic memory (\textsc{pc}) operates on DINOv2 tokens, whereas the textural memory (\textsc{pc-cnn}) uses Wide-ResNet50~\cite{zagoruyko2016wrn} layer-2/3 features. These views are deliberately complementary: CNN features preserve fine local textures that semantic ViT tokens tend to smooth out, while ViT features provide stronger semantic consistency (Sec.~\ref{sec:ablation}).

Together, the set-feature and dual-memory representations provide complementary appearance evidence, capturing both composition-level deviations and fine-grained structural defects.

\subsection{Training-free open-vocabulary counting}
\label{sec:count}

The appearance representations are insensitive to changes in the number of identical parts, which is precisely what many logical anomalies alter. We therefore introduce an explicit training-free counting cue. Concretely, we employ a frozen promptable concept-segmentation model (SAM3) purely as a counting primitive.

Given a small set of open-vocabulary concept names, $\{q_1,\dots,q_C\}$
such as ``screw'', ``nut'', and ``washer'', the frozen segmenter returns confidence-scored instance proposals without any category-specific training. We count an instance whenever its confidence exceeds a threshold. However, counting at a single confidence threshold is sensitive to the choice of operating point. Rather than committing to one threshold, we repeat the counting process across $T$ confidence levels and concatenate the resulting counts into a count vector $c\in\mathbb{R}^{D}$ with $D=CT$. This multi-threshold representation reveals which concepts are counted consistently across operating points and therefore provides a more stable characterization than any single-threshold estimate.

\noindent \textit{Per-dimension reliability.} Not all count dimensions are equally reliable. Concepts whose counts fluctuate substantially on normal images (e.g., granola flakes) should contribute less to anomaly scoring than stable concepts. We quantify this reliability using the held-out normal validation set $\mathcal{N}_{\mathrm{val}}$. For each dimension $j$, we estimate its mean $\mu_j$ and standard deviation $\sigma_j$, with $\sigma_j$ lower-bounded by $\tau$, and compute the coefficient of variation, $v_j=\sigma_j/(|\mu_j|+\tau)$. We then define a reliability weight $w_j=1/(1+(v_j/v_0)^2)$, so that dimensions exhibiting stable normal counts automatically receive larger weights.

Finally, the counting anomaly score is computed as the reliability-weighted standardized deviation, $s_{\mathrm{sam}}=\sum_j w_j\,|c_j-\mu_j|/\sigma_j$. This procedure turns a frozen open-vocabulary segmenter into a robust training-free counter, providing the explicit counting signal that appearance representations lack.

\begin{figure}[t]
\centering
\includegraphics[width=0.9\textwidth]{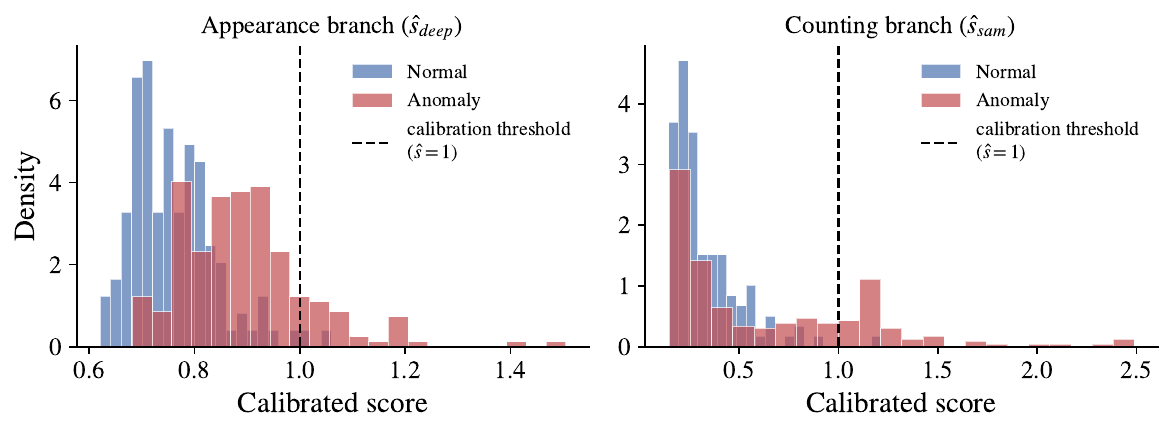}
\caption{
Normal-set calibration renders heterogeneous branch scores directly comparable. On the \textit{screw\_bag} category, calibrated appearance (\textsc{deep}) and counting (\textsc{sam}) scores place normal images near the shared threshold ($\hat{s}=1$, dashed), while anomalies extend beyond it in both branches. This branch-independent scale makes heterogeneous cues directly fusible.
}
\label{fig:calib}
\end{figure}

\subsection{Normal-set calibration for heterogeneous cues}
\label{sec:calib}

The branches provide complementary evidence but produce heterogeneous score distributions, making naive fusion unreliable. Our key observation is that each branch can define its own notion of normality using only normal data. For each branch $b$, we estimate the $99$th percentile of its scores over the held-out normal validation set, $Q_b=Q_{0.99}\!\left(s_b(\mathcal{N}_{\text{val}})\right)$
and calibrate the raw score as
\begin{equation}
\hat{s}_b = s_b / Q_b,
\end{equation}

This maps the normal $99$th percentile of every branch to a shared threshold at $\hat{s}_b=1$, so that normal images typically satisfy $\hat{s}_b\!\lesssim\!1$, whereas stronger deviations extend beyond it (Fig.~\ref{fig:calib}). Using only unlabeled normal images, normal-set calibration transforms otherwise incomparable scores into commensurable ones and serves as the key enabler of training-free fusion. All subsequent interactions between branches operate exclusively on the calibrated scores $\hat{s}_b$.

\subsection{Soft-OR fusion}
\label{sec:fusion}

Complementary branches frequently provide sparse evidence: one branch may respond strongly while others remain close to normal, as confirmed by the oracle analysis in Sec.~\ref{sec:ablation}. A plain sum dilutes such evidence among silent branches, whereas a max is overly sensitive to single-branch false positives. Because the calibrated scores are non-negative, $\hat{s}_b\ge0$, We instead aggregate them using a $p$-norm,
\begin{equation}
S=\Big(\textstyle\sum_{b\in\mathcal{B}}\hat{s}_b^{\,p}\Big)^{1/p},
\end{equation}

The $p$-norm interpolates between the sum ($p\!=\!1$) and the max ($p\rightarrow\infty$). In practice, we use $p\!=\!6$ (validated in Sec.~\ref{sec:ablation}), which emphasizes strong evidence without allowing a single outlier to dominate. The resulting fusion therefore behaves as a soft logical OR over complementary branches.

\subsection{Interpretable evidence attribution}
\label{sec:explain}

Because calibration defines a shared notion of ``above normal,'' the same scores used for detection can also support explanation. For each branch, we define an above-normal trigger $t_b=\max(\hat{s}_b-1,0)$, which activates only when the calibrated score exceeds its normal threshold.

We aggregate these triggers over a logical axis, $\textsc{deep},\textsc{pix},\textsc{sam}$
and a structural axis, $\textsc{pc},\textsc{pc-cnn}$. The dominant axis provides a coarse logical versus structural attribution for the detection. For logical detections, the count dimension with the largest reliability-weighted contribution identifies the offending concept and reports its observed versus normal count (e.g., ``label: $0$ vs.\ $2$''). The associated instance masks localize the supporting evidence. Structural detections are localized using the patch-memory distance maps. Importantly, this attribution reuses the detector's own calibrated evidence; no additional model is trained for explanation.

\subsection{Setup}

\textit{Dataset and metric.}
We evaluate on MVTec-LOCO~\cite{bergmann2022loco}, reporting image-level AUROC separately on logical and structural anomalies and their average. We additionally evaluate on MVTec-AD~\cite{bergmann2019mvtec}, reporting mean image-AUROC to assess structural generalization beyond logical anomalies (Sec.~\ref{sec:generalisation}).

\noindent\textit{Implementation.}
All models remain frozen throughout. The set-feature branches and semantic memory use DINOv2 ViT-L features (four intermediate layers for set descriptors and the final layer for memory), the textural memory uses Wide-ResNet50 layer-2/3 features, and counting employs SAM3. Importantly, the entire protocol remains training-free. Patch memories and set-feature references are built from normal training images, while calibration quantiles are estimated from the held-out normal validation set. No anomaly labels, part annotations, gradient optimization, LLMs, or category-specific networks are used. Unless otherwise stated, all hyperparameters are shared across categories: set descriptors use 1,000 projections, patch memories retain a 10\% coreset with top-1\% scoring, counts aggregate over $T=5$ thresholds, and $(p,v_0,\tau)=(6,0.25,0.3)$. The only category-specific input is a small set of generic concept names for counting.

Inference requires only frozen forward passes on a single GPU, taking roughly $0.3$--$1.2$\,s per image depending on the number of concepts (83\,ms for DINOv2+Wide-ResNet50 and 0.27\,s per concept for counting).

\begin{table}[t]
\centering
\caption{
Comparison with prior work on MVTec-LOCO (image-level AUROC, \%). Methods are grouped by the supervision they require. Best overall results are shown in \textbf{bold}, and the best training-free results are \underline{underlined}. PatchCore results are from~\cite{hsieh2024csad}; all others are taken from the cited papers.
}
\label{tab:sota}
\begin{tabular}{llccc}
\toprule
& Method & Logical & Structural & Average \\
\midrule
\multirow{1}{*}{\rotatebox{90}{labels}}
& PSAD$^{\dagger}$~\cite{kim2024psad}            & \textbf{98.1} & 91.6 & \textbf{94.9} \\
\midrule
\multirow{5}{*}{\rotatebox{90}{trained}}
& GCAD~\cite{bergmann2022loco}                    & 86.0 & 80.7 & 83.4 \\
& EfficientAD-M~\cite{batzner2024efficientad}     & 86.8 & 94.7 & 90.7 \\
& SLSG~\cite{yang2023slsg}                        & 89.6 & 91.4 & 90.3 \\
& PUAD~\cite{sugawara2024puad}                    & 92.0 & 94.1 & 93.1 \\
& CSAD~\cite{hsieh2024csad}                       & 96.7 & 94.0 & 95.3 \\
\midrule
\multirow{4}{*}{\rotatebox{90}{training-free}}
& PatchCore~\cite{roth2022patchcore}             & 79.7 & 87.7 & 83.7 \\
& ComAD+PatchCore~\cite{liu2023comad}            & 89.4 & 90.9 & 90.1 \\
& SINBAD~\cite{cohen2023sinbad}                  & \underline{91.2} & 85.5 & 88.3 \\
& \textbf{Ours}                                   & 89.0 & \textbf{\underline{95.9}} & \underline{\textbf{92.5}} \\
\bottomrule
\end{tabular}
\end{table}

\subsection{Comparison with prior work}

Table~\ref{tab:sota} compares our method with published results on MVTec-LOCO, grouped by the supervision each method requires. Among training-free methods, our detector achieves the highest average image-AUROC ($92.5$), surpassing SINBAD ($88.3$) and ComAD+PatchCore ($90.1$). This advantage is largely driven by structural anomalies, where our score of $95.9$ is the strongest in the comparison, exceeding trained detectors such as EfficientAD ($94.7$) and PUAD ($94.1$).

The remaining gap lies primarily in logical anomalies that benefit from clean part representations. Our logical AUROC ($89.0$) remains below methods that explicitly learn category-specific part models, such as CSAD and the part-supervised PSAD, and trails SINBAD among training-free approaches. This behaviour is consistent with the stronger supervision available to these methods: segmentation-based detectors exploit clean per-category part representations, whereas our counting branch relies solely on a frozen open-vocabulary segmenter without training or annotations.

Table~\ref{tab:percat} further breaks down the logical results by category and reveals substantial variation across anomaly types. Our method is particularly effective when count or composition cues are relatively distinct, achieving the strongest training-free result on breakfast\_box ($99.4$) and approaching the part-supervised PSAD. The most challenging category is screw\_bag ($77.2$), where numerous small, near-identical parts make reliable counting difficult. This category accounts for much of the remaining gap to part-supervised approaches and highlights the limitations of the current counting primitive.

\begin{table}[t]
\centering
\caption{
Per-category breakdown of logical anomaly detection on MVTec-LOCO (AUROC, \%). Best results in each category are shown in \textbf{bold}. $^\dagger$ denotes part supervision. Results are taken from the cited papers.
}
\label{tab:percat}
\begin{tabular}{lcccc}
\toprule
Category & GCAD & SINBAD & PSAD$^{\dagger}$ & Ours \\
\midrule
breakfast\_box       & 87.0  & 97.7          & \textbf{100.0} & 99.4 \\
juice\_bottle        & \textbf{100.0} & 97.1 & 99.1           & 92.7 \\
pushpins             & 97.5  & 88.9          & \textbf{100.0} & 89.2 \\
screw\_bag           & 56.0  & 81.1          & \textbf{99.3}  & 77.2 \\
splicing\_connectors & 89.7  & 91.5          & \textbf{91.9}  & 86.5 \\
\midrule
\textbf{Average}     & 86.0  & 91.2          & \textbf{98.1}  & 89.0 \\
\bottomrule
\end{tabular}
\end{table}

\begin{figure}[t]
\centering
\begin{minipage}[t]{0.55\textwidth}\centering
\includegraphics[width=\textwidth]{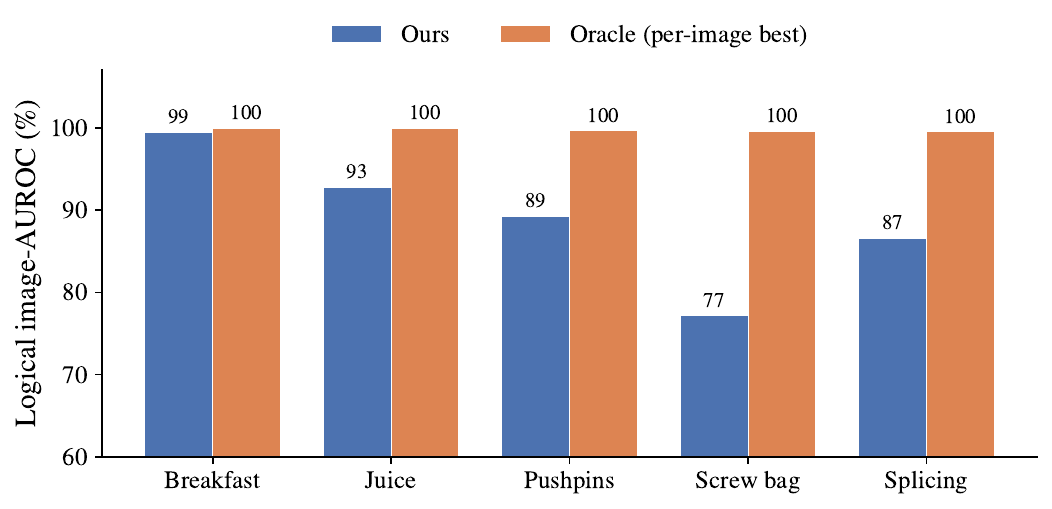}
\caption{
Oracle upper bound for logical anomaly detection (image-AUROC, \%). For each image, the oracle selects the calibrated branch that best separates it from normal samples. Because this choice requires ground-truth labels, it serves only as an unattainable upper bound.
}
\label{fig:oracle}
\end{minipage}\hfill
\begin{minipage}[t]{0.35\textwidth}\centering
\includegraphics[width=\textwidth]{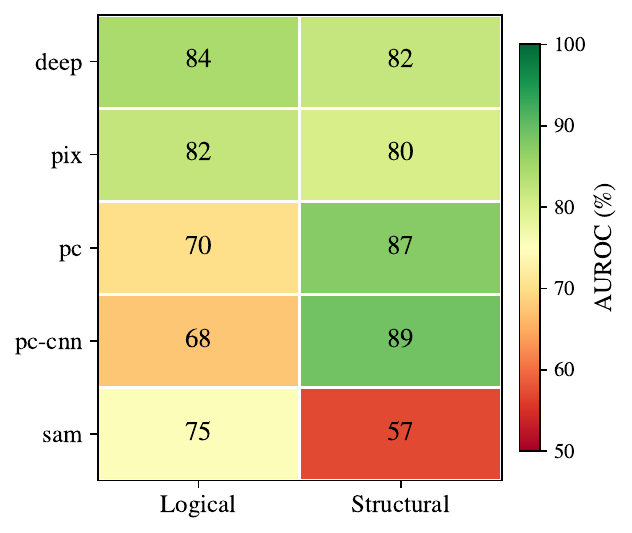}
\caption{
Logical versus structural specialization of the branches. Each cell reports a branch's average image-AUROC on logical or structural anomalies. Branch definitions follow Sec.~\ref{sec:method}.
}
\label{fig:branch}
\end{minipage}
\end{figure}

\begin{figure}[t]
\centering
\includegraphics[width=\textwidth]{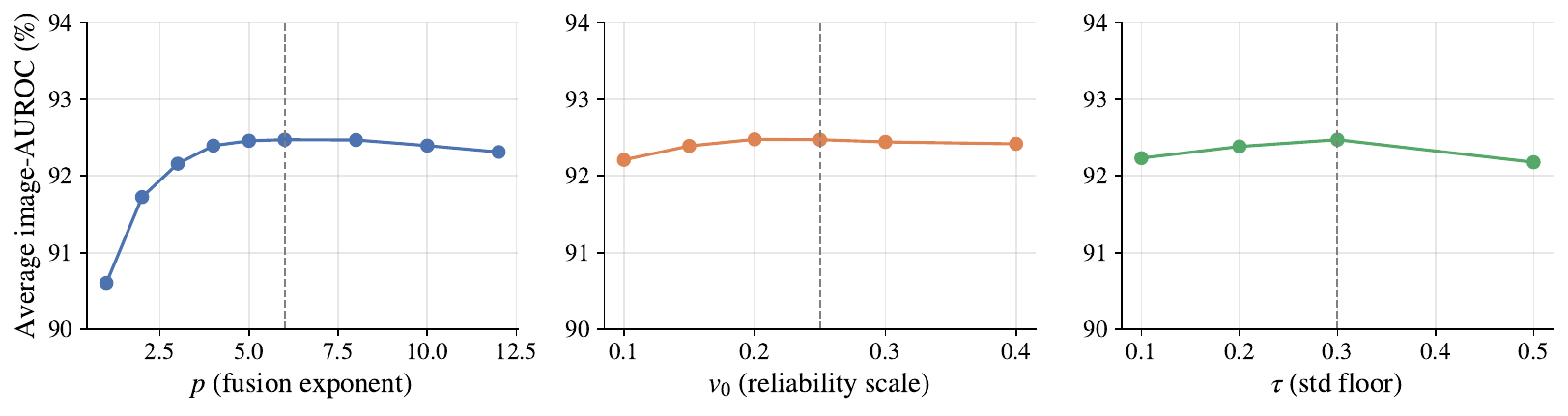}
\caption{
Sensitivity to the shared hyperparameters on MVTec-LOCO (average image-AUROC, \%). Dashed lines indicate the values used in all experiments. Performance remains stable across broad ranges of the fusion exponent $p$, reliability scale $v_0$, and standard-deviation floor $\tau$.
}
\label{fig:sens}
\end{figure}

\subsection{Ablation and Diagnostic Analysis}\label{sec:ablation}

\noindent\textit{Complementarity of the branches.}
Leave-one-out analysis (Table~\ref{tab:abl}, left) shows that the branches contribute to different anomaly types rather than redundantly modelling the same signal. Removing the counting branch reduces logical AUROC by $3.4$ points, whereas removing the textural memory decreases structural AUROC by $1.6$ points. The semantic memory is the strongest individual contributor on the structural axis ($-4.2$ points). This limited overlap explains why the branches combine constructively.

\noindent\textit{Fusion as the remaining bottleneck.}
Figure~\ref{fig:oracle} shows that a per-image oracle selecting the best calibrated branch for each image reaches nearly $100$ AUROC on both anomaly axes in every category. Since the oracle differs only in how branch outputs are combined, this result suggests that most of the required signal is already present and that the remaining limitation lies in fusion. Figure~\ref{fig:branch} supports this interpretation: the appearance branches act as generalists, the patch memories specialise in structural anomalies, and the counting branch specialises in logical anomalies.

\noindent\textit{Fusion strategy.}
Given this complementarity, the aggregation rule determines how much of the available signal can be retained in a single score. Table~\ref{tab:abl} (right) shows that the calibrated $p$-norm achieves the strongest average performance ($92.5$), outperforming the calibrated sum ($90.6$), rank-max ($89.0$), and top-2 rank pooling ($89.7$). The sum dilutes single-branch evidence, whereas the max overreacts to isolated outliers; the $p$-norm balances the two.

\noindent\textit{Robustness and counting design.}
Figure~\ref{fig:sens} shows broad performance plateaus across the shared hyperparameters, indicating that the chosen settings are not overly sensitive. Reliability weighting is essential for the counting branch: a plain Mahalanobis distance over the count vector performs only marginally above chance, whereas the reliability-weighted deviation recovers strong logical performance. Curated concept names further outperform enlarged generic prompt sets, suggesting that lightweight semantic guidance remains beneficial for counting.

\begin{table}[t]
\centering
\caption{Ablations (MVTec-LOCO average image-AUROC). Left: leave-one-branch-out from the full model. Right: fusion rule at fixed branches. Best average in \textbf{bold}.}
\label{tab:abl}
\small
\setlength{\tabcolsep}{4pt}
\begin{tabular}{lccc@{\hskip 0.6cm}lc}
\toprule
Variant & Logical & Structural & Average & Fusion rule & Average \\
\midrule
Ours (full)   & 89.0 & 95.9 & \textbf{92.5} & $p$-norm ($p\!=\!6$) & \textbf{92.5} \\
$-$\,counting & 85.6 & 96.7 & 91.1 & plain sum      & 90.6 \\
$-$\,textural & 88.7 & 94.3 & 91.5 & rank-max       & 89.0 \\
$-$\,semantic & 89.9 & 91.7 & 90.8 & top-2 pool     & 89.7 \\
\bottomrule
\end{tabular}
\end{table}

\subsection{Evaluating the Explanations}

The explanations are largely faithful to the detector's decisions. Using the decoding of Sec.~\ref{sec:explain}, the predicted logical/structural attribution agrees with the ground-truth anomaly type for $73.8\%$ of anomalies on average (Table~\ref{tab:explain}), indicating that the explanation reflects the evidence driving the detector.

\begin{figure}[tbp]
\centering
\includegraphics[width=0.9\textwidth]{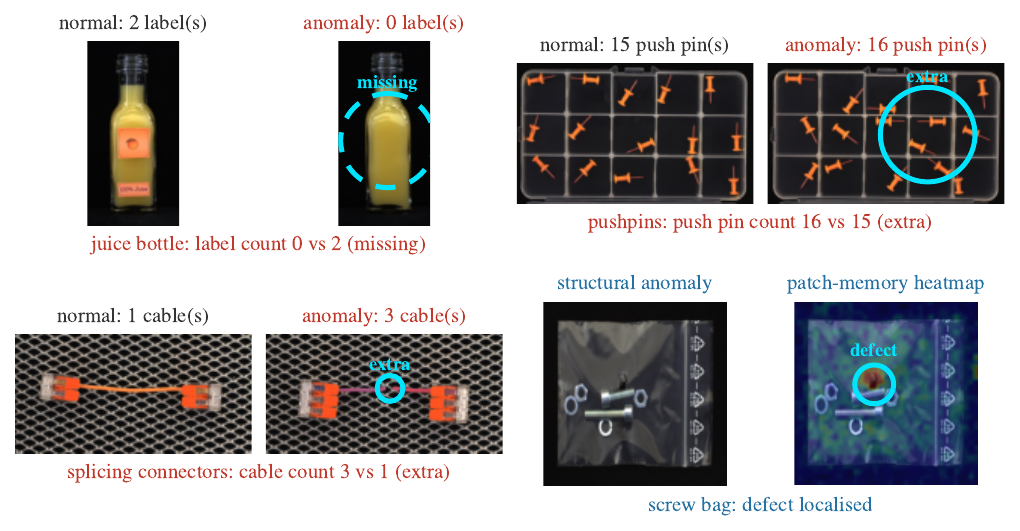}
\caption{
Self-explaining outputs for logical and structural anomalies. Logical anomalies are explained through detected concept instances and their count rationales, whereas structural anomalies are localised by the patch-memory heatmap. Cyan circles mark the labelled anomalous regions for reference (dashed when the object is missing). All explanations reuse the calibrated scores used for detection and axis attribution, requiring no additional explanation model.
}
\label{fig:explain}
\end{figure}

The explanations are also human-readable. For logical anomalies, the dominant reliability-weighted count dimension identifies the offending concept and reports its observed versus normal count, while the corresponding instance masks localise the evidence. Structural anomalies inherit the localisation behaviour of the patch memories through their distance maps. Representative examples are shown in Fig.~\ref{fig:explain}. Importantly, these explanations are intrinsic to the detector: they reuse the same calibrated scores employed for detection and fusion, without training an additional explanation model. Unlike most training-free detectors, which typically provide only anomaly scores or heatmaps, our method additionally reports a count rationale tied directly to its prediction.
\begin{table}[t]
\centering
\caption{Explainability: accuracy of the logical/structural axis attribution per category (\% of anomalies whose predicted axis matches the ground-truth type).}
\label{tab:explain}
\begin{tabular}{lcccccc}
\toprule
 & breakfast & juice & pushpins & screw\_bag & splicing & \textbf{Average} \\
\midrule
Axis-attr.\ acc. & 81.5 & 78.0 & 52.9 & 81.3 & 75.1 & \textbf{73.8} \\
\bottomrule
\end{tabular}
\end{table}

\subsection{Structural Generalisation}\label{sec:generalisation}

To assess whether the structural subsystem generalises beyond MVTec-LOCO, we evaluate it on the 15 categories of MVTec-AD~\cite{bergmann2019mvtec}, which contains only structural anomalies. We therefore retain the two calibrated patch memories and the same $p$-norm fusion while disabling the logical branches.

Table~\ref{tab:mvtecad} summarises the results. The structural configuration achieves an average image-AUROC of $99.1$, matching PatchCore~\cite{roth2022patchcore}. In contrast, applying the full five-branch LOCO configuration markedly reduces performance ($78.4$ average), indicating that the logical branches are specialised to compositional anomalies rather than universally beneficial. This behaviour confirms that the proposed modular design transfers to standard structural anomaly detection when configured to the anomaly types present.

\begin{table}[t]
\centering
\caption{
Structural generalisation to MVTec-AD (per-category image-AUROC, \%). Because MVTec-AD contains only structural anomalies, we evaluate the structural configuration of our framework while disabling the logical branches.
}
\label{tab:mvtecad}
\begin{tabular}{lc|lc|lc}
\toprule
Category & AUROC & Category & AUROC & Category & AUROC \\
\midrule
bottle  & 100.0 & hazelnut  & 100.0 & tile       & 100.0 \\
cable   & 95.3  & leather   & 100.0 & toothbrush & 96.9  \\
capsule & 99.2  & metal\_nut & 100.0 & transistor & 100.0 \\
carpet  & 99.9  & pill      & 98.6  & wood       & 99.8  \\
grid    & 100.0 & screw     & 96.5  & zipper     & 100.0 \\
\midrule
\multicolumn{6}{c}{\textbf{Average $99.1$} \quad (DINOv2 memory only: $98.6$; full five-branch fusion: $78.4$)} \\
\bottomrule
\end{tabular}
\end{table}

\section{Discussion}

Three broader lessons emerge from our analysis. First, the oracle study suggests that the bottleneck is not signal acquisition but signal integration: a per-image oracle over the branches reaches nearly $100$ AUROC on both anomaly axes (Fig.~\ref{fig:oracle}), implying that the necessary evidence is already present. The challenge is therefore to integrate heterogeneous cues without suppressing single-branch evidence. Second, normal-set calibration provides the common language that makes such fusion possible. By mapping branch scores to a shared normal range using only normal images, it enables out-of-distribution cues such as open-vocabulary counting to be fused with appearance-based detectors without training. Third, the framework is naturally modular. The same calibrated $p$-norm supports both the full logical--structural detector on MVTec-LOCO and a competitive structural detector on MVTec-AD when the logical branches are disabled.

\begin{figure}[tbp]
\centering
\includegraphics[width=0.75\textwidth]{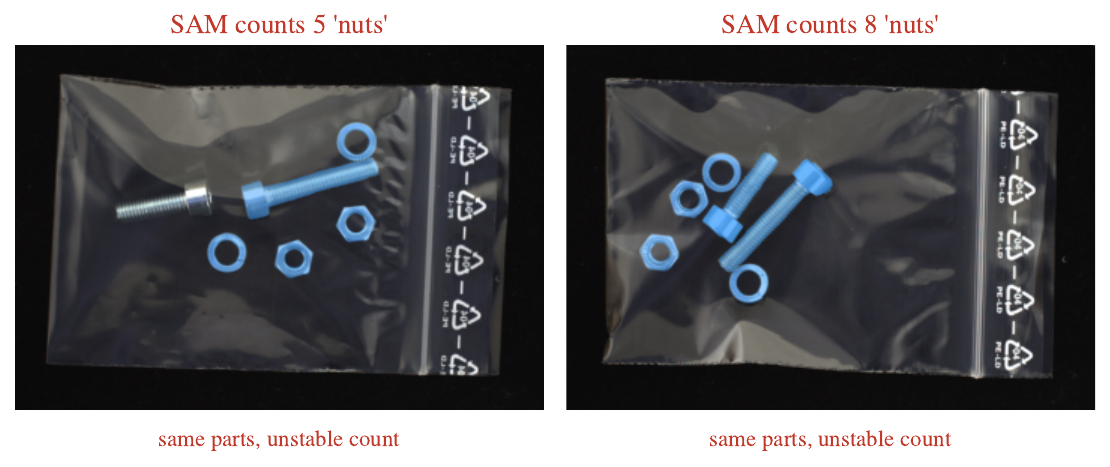}
\caption{
Failure case: unstable counting on \textit{screw\_bag}. Two normal images with identical parts receive different "nut" counts from the frozen segmenter, reflecting its difficulty in separating small, visually similar components. The resulting variability broadens the normal count distribution and limits logical discrimination on this category.
}
\label{fig:failure}
\end{figure}

\noindent\textit{Limitations.} Broader evaluation of logical anomalies remains constrained by the scarcity of public benchmarks beyond MVTec-LOCO. The counting branch is limited by the frozen segmenter, which struggles with small, cluttered, visually similar parts and broadens the normal count distribution on categories such as screw\_bag (Fig.~\ref{fig:failure}). The method also requires a small number of category concept names, as a fully name-free variant recovers only about half of the counting gain (Sec.~\ref{sec:ablation}). Finally, the wide oracle gap indicates that training-free single-score fusion still cannot fully exploit branch complementarity, leaving fusion itself as the clearest opportunity for future improvement.

\section{Conclusion}
This work addressed the challenge of training-free detection across both logical and structural anomalies. Our key insight is that heterogeneous frozen-model cues can be brought onto a common scale through normal-set calibration and fused without training, allowing appearance-based evidence and open-vocabulary counting to complement one another in a single self-explaining detector. The resulting framework is competitive across both anomaly families, achieving the strongest average performance among training-free methods on MVTec-LOCO while remaining a competitive structural detector on MVTec-AD. The remaining challenge is not obtaining additional cues but integrating complementary cues more effectively, as highlighted by the wide oracle gap.

\bibliographystyle{splncs04}
\bibliography{references}
\end{document}